\documentclass[sigconf,natbib=false]{acmart}

\usepackage{graphicx}
\usepackage{dcolumn}
\usepackage{bm}
\usepackage{color}
\usepackage{hyperref} 
\usepackage{tabularx} 
\usepackage{amsmath}
\usepackage{multirow}
\usepackage{subcaption}
\usepackage{float}
\usepackage{booktabs} 
\usepackage{url}
\usepackage{xcolor} 

\newcolumntype{Y}{>{\raggedright\arraybackslash}X} 
\newcolumntype{C}{>{\centering\arraybackslash}X}   

\AtBeginDocument{%
  }

\setcopyright{acmlicensed}
\copyrightyear{2024}
\acmYear{2024}
\acmDOI{}

\acmISBN{}

\RequirePackage[
  datamodel=acmdatamodel,
  style=acmnumeric,
  ]{biblatex}

\begin{document}

\title{Enhancing Large Language Models with Domain-Specific Knowledge: The Case in Topological Materials}


\author{Huangchao Xu}
\affiliation{%
  \institution{Computer Network Information Center, Chinese Academy of Sciences}
  \country{}
}
\affiliation{
    \institution{University of Chinese Academy of Sciences}
    \state{Beijing}
    \country{China}
}
\email{hcxu@cnic.cn}

\author{Baohua Zhang}
\authornotemark[1]
\affiliation{%
  \institution{Computer Network Information Center, Chinese Academy of Sciences}
  \state{Beijing}
  \country{China}
  }
\email{zhangbh@sccas.cn}

\author{Zhong Jin}
\affiliation{%
  \institution{Computer Network Information Center, Chinese Academy of Sciences}
  \state{Beijing}
  \country{China}
  }
\email{zjin@sccas.cn}

\author{Tiannian Zhu}
\affiliation{%
  \institution{Institute of Physics, Chinese Academy of Sciences}
  \state{Beijing}
  \country{China}
  }
\email{zhutiannian15@mails.ucas.ac.cn}

\author{Quansheng Wu}
\authornotemark[1]
\affiliation{%
  \institution{Institute of Physics, Chinese Academy of Sciences}
  \state{Beijing}
  \country{China}
  }
\email{quansheng.wu@iphy.ac.cn}

\author{Hongming Weng}
\affiliation{%
  \institution{Institute of Physics, Chinese Academy of Sciences}
  \state{Beijing}
  \country{China}
  }
\email{hmweng@iphy.ac.cn}

\renewcommand{\shortauthors}{Huangchao Xu et al.}

\begin{abstract}
Large language models (LLMs), such as ChatGPT, have demonstrated impressive performance in the text generation task, showing the ability to understand and respond to complex instructions. However, the performance of naive LLMs in specific domains is limited due to the scarcity of domain-specific corpora and specialized training. Moreover, training a specialized large-scale model necessitates significant hardware resources, which restricts researchers from leveraging such models to drive advances. Hence, it is crucial to further improve and optimize LLMs to meet specific domain demands and enhance their scalability. Based on the condensed matter data center, we establish a material knowledge graph (MaterialsKG) and integrate it with literature. Using large language models and prompt learning, we develop a specialized dialogue system for topological materials called TopoChat. Compared to naive LLMs, TopoChat exhibits superior performance in structural and property querying, material recommendation, and complex relational reasoning. This system enables efficient and precise retrieval of information and facilitates knowledge interaction, thereby encouraging the advancement on the field of condensed matter materials.
\end{abstract}




\maketitle

\section{Introduction}
Large language models (LLMs) represent the latest significant advancements in the field of natural language processing. They can not only generate natural language text, but also possess a profound comprehension of textual semantics and perform various tasks such as summarizing text, answering questions, translating languages, and more. ChatGPT\cite{RAY2023121} and GPT-4\cite{bubeck2023sparks} are pre-trained on massive datasets and boast enormous parameter sizes, reaching 175 billion and 1.8 trillion, respectively. Through techniques such as pretraining, fine-tuning, and reinforcement learning with human feedback, These models exhibit a remarkable capacity to precisely understand and effectively tackle complex inquiries. Meta has also introduced the LLama series\cite{touvron2023llama} of pre-trained large language models, offering a range of versions with parameter sizes spanning from 7 billion to 70 billion for LLama2 and LLama3 at present. In April, Qwen has made a significant contribution to the field by open-sourcing Qwen1.5-110B\cite{qwen}, which boasts an expansive parameter count of one hundred billion. At the same time, Zhipu AI has unveiled their latest models ChatGLM4(the upgraded version of ChatGLM3\cite{du2022glm}\cite{zeng2022glm}), whose capabilities are gradually approaching GPT-4. The emergence of the universal LLMs provides a powerful tool for the scientific field and gives researchers more potential capabilities.

Despite the notable achievements of LLMs in various aspects, they are with some limitations. One prominent drawback is the occurrence of hallucinations, in which the model may generate false or inaccurate information. Furthermore, the application of LLMs in professional domains encounters greater challenge due to the scarcity of domain-specific training corpora, impeding the development of specialized fields. In addition, retraining of large language models imposes significant demands in terms of data and hardware resources, necessitating abundant high-quality training data and computational capabilities. Considering that current LLMs are trained on extensive general-purpose corpora, capable of fulfilling diverse natural language processing requirements, a prevailing trend in both research and application is to leverage LLMs as foundational models and incorporate domain-specific data to achieve a balance between generality and specialization. This approach enables the integration of specialized knowledge and effectively addresses the limitations of LLMs, enhancing accuracy and reliability in scientific domains.

In the field of condensed matter science, researchers traditionally utilize machine learning or deep learning techniques to discover new materials or properties. Xu\cite{PhysRevB.109.035122} presented a study using supervised machine learning to train neural networks, developing a heuristic chemical rule that accurately diagnoses whether a material is topological. Liu et al\cite{liu2023materials} used a Gaussian process regression model with a specialized kernel to obtain effective descriptors for predicting topological semimetals among square-net materials. Graph neural networks are often used for crystal characterisation and training among deep learning techniques. Lee and Xia\cite{lee2024machine} showed that they use a graph neural network-based machine learning universal harmonic interatomic potential (MLUHIP) to accurately predict phonon frequencies and thermodymanic properties for unseen crystalline compounds. Ashiqur Rasul and colleagues\cite{rasul2023topological} demonstrated an approach that integrates a graph convolutional neural network with atom-specific persistent homology for the purpose of accurately classifying materials as either topological or trivial.

In recent years, researchers also employ pre-trained or fine-tuned models for the investigation of material properties. Das et al introduced CrysGNN\cite{das2023crysgnn}, a new pre-trained graph neural network (GNN) framework designed to enhance property prediction for crystalline materials. Dong et al\cite{dong2023discovery} showed that they use a pipeline combining a transformer-based composition generator, template-based structure prediction, and graph neural network-based relaxation to obtain the discovery of new stable hypothetical 2D materials. Huang et al\cite{huang2023materials} developed the Materials Informatics Transformer (MatInFormer) which learns the grammar of crystallography through tokenization of space group information, for material property prediction. Rubungo et al\cite{rubungo2023llmprop} utilized a method called LLM-Prop to study the prediction of physical and electronic properties of crystalline solids from their text descriptions. Song et al\cite{song2023bridging} utilized a numerical reasoning method for material Knowledge Graphs (NR-KG) to study material property prediction. While current research has yielded certain accomplishments, there remains a necessity to explore artificial intelligence methods that focus on a more expansive framework.

The Condensed Matter Physics Data Center (CMPDC, \href{https://cmpdc.iphy.ac.cn/}{\url{https://cmpdc.iphy.ac.cn/}}) serves as a comprehensive platform for diverse scientific data generated by prominent researchers in the field\cite{Zhang_2019}. It offers over 17 featured databases that encompass a wide range of theoretical calculations and experimental measurements in condensed matter physics. These resources cover crucial aspects of research, including analysis of component structures, characterization and measurement techniques, property modulation methodologies, effect detection strategies. The databases provide a solid foundation for the construction of intelligent and specialized models in condensed matter science. However, several challenges persist for material science based on current data center. Firstly, despite the abundance of diverse data, the issue of data silos remains a significant concern. The heterogeneity of data sources, formats, and standards requires an effective fusion strategy to establish a unified and high quality database that serves as the foundation for building large-scale models. Secondly, the construction of a material expert system poses great challenges. Such a system should provide comprehensible and accurate domain knowledge while enabling complex inference. Currently, users are constrained to structured query methods provided by separated databases, which offer limited query results and lack the ability to handle complex query logic. Furthermore, it is worth noting that the query results obtained from current databases often demand a substantial level of domain knowledge to understand. As a consequence, the development of an expert system equipped with natural language interaction becomes necessary. 

The aim of this paper is to effectively address the mentioned challenges by integrating multi-source knowledge into large language models and proposing an expert system named TopoChat. To illustrate the implementation of this system, topological material is taken as a representative example. Our contributions in this paper can be outlined as follows:
\begin{itemize}
    \item We develop a multi-source material data management flow that enables seamless integration of diverse sources. By leveraging this flow, we achieve a comprehensive and unified data representation, which significantly enhances the analysis and understanding of topological materials.
    \item We construct a powerful material knowledge graph (MaterialsKG) and implement a literature extraction fission strategy to enhance large language models. 
    \item We develop TopoChat dialogue system tailored specifically for topological materials. By following several guidelines for prompts, TopoChat has shown superior performance compared to base large models in answering various material-related questions.
\end{itemize}

\section{Methodology}
\subsection{Data Preparation}
CMPDC integrates data from multiple sites, incorporating a wide range of data sources. This includes a substantial collection of historical data generated through experiments and theoretical calculations. The data within portal exhibits a relatively decentralized distribution, with various formats and standards across multiple databases, accommodating both relational and non-relational data structures. For example, The topological materials database comprises detailed topological classifications of over 28,000 materials,  derived from first-principles calculations and theoretical derivations\cite{He_2019}. The crystal structure and diffraction database includes a large amount of crystal diffraction data, surpassing 400,000 entries, obtained through theoretical derivations based on experimental crystal structures. The material phononics database has a rich collection of more than 10,000 entries, containing topological phonon properties acquired through high-throughput calculations on real materials\cite{li2021computation}. The inter connectivity between these distinct databases primarily relies on different interfaces, thereby posing challenges in providing comprehensive support for diverse data analyses and applications. 

\begin{figure}[h]
  \centering
  \includegraphics[width=0.5\textwidth]{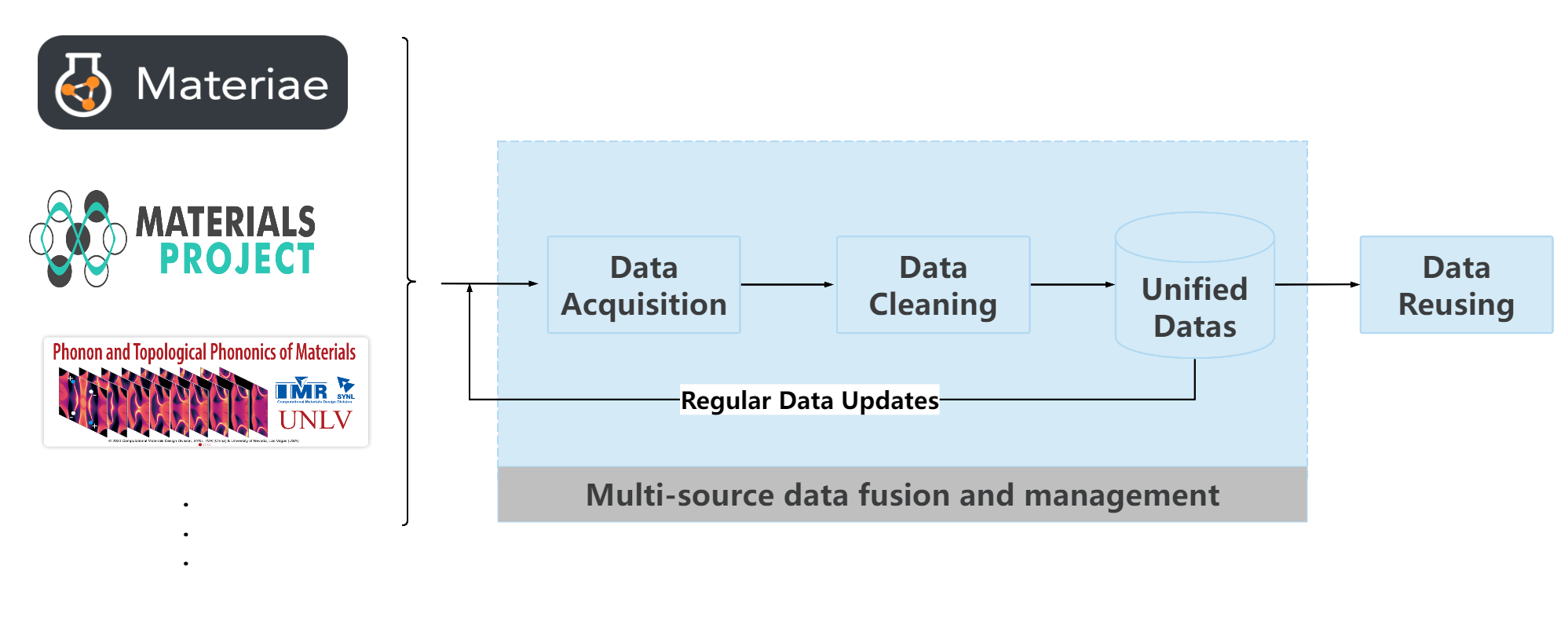}
  \caption{Multi-Source Data Management Flow}
  \label{fig:1}
\end{figure}

Multi-source data fusion and management flow is aimed at bridging the data barriers between sites to build a unified high-quality database that enables seamless data connectivity between different sites, as shown in Figure \ref{fig:1}, where the process includes data acquisition, data cleaning, data storage, data maintenance and reuse. We utilize the pymatgen\cite{ONG2013314} and mp\_api tool provided by the Materials Project to gather essential information, such as cell parameters, density based on the material id. This information is combined with the energy gap, indirect energy gap, and fermi energy provided by CMPDC. Additionally, we incorporate topological phonon properties to create a more extensive description of material attributes. There are seven crystal systems: triclinic, monoclinic, tetragonal, cubic, orthorhombic, hexagonal, and trigonal. These crystal systems are subdivided into 230 space groups and 32 different point groups, which are consistently denoted by schoenflies' symbol. The topological classes can be classified into six categories: topological insulator, topological crystalline insulator, high-symmetry line semimetal, high-symmetry point semimetal, generic-momenta semimetal, and trivial insulator. This taxonomy was proposed by the Institute of Physics, Chinese Academy of Sciences in 2019\cite{Zhang_2019}..

\begin{figure}[h]
  \centering
  \includegraphics[width=0.5\textwidth]{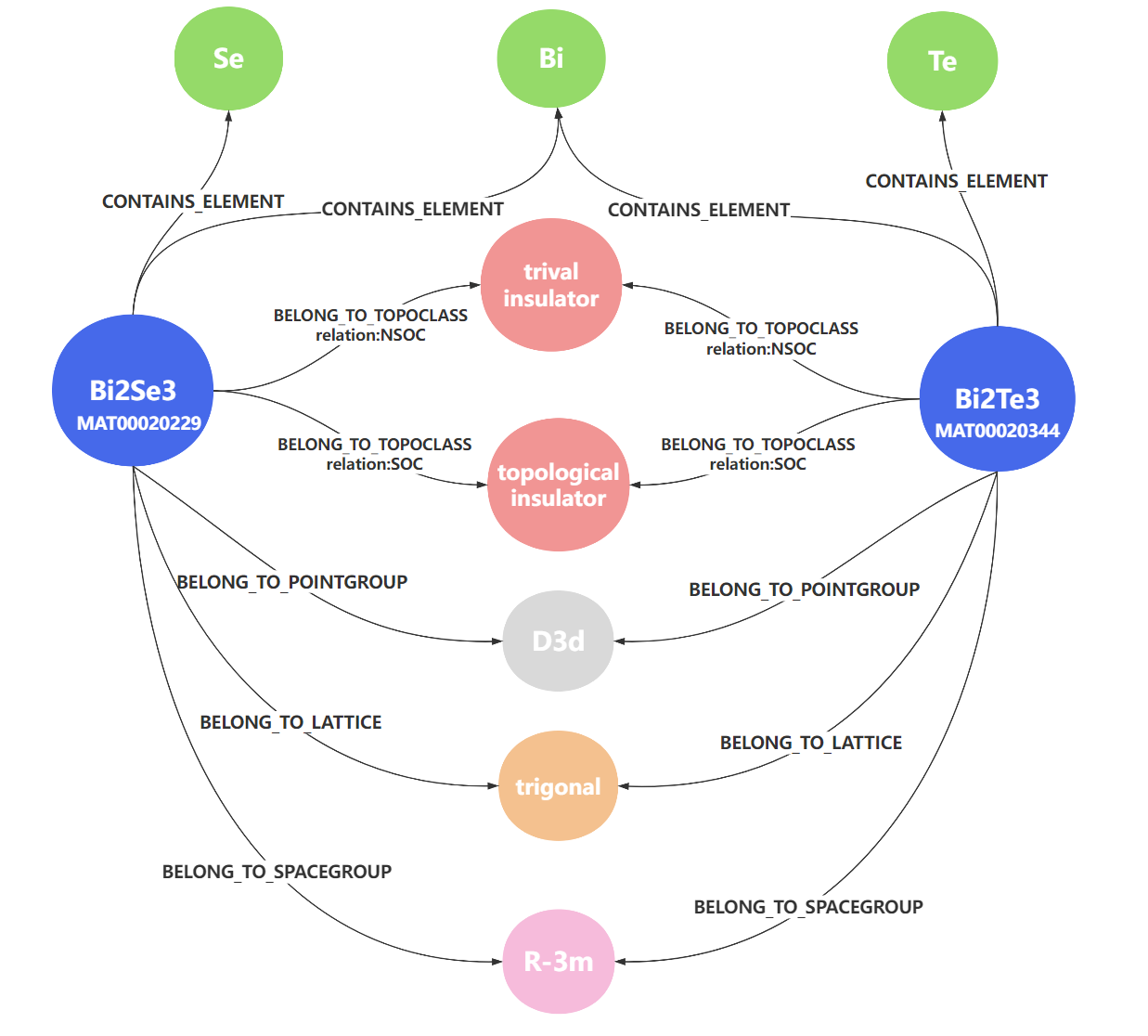}
  \caption{The Components of Materials Knowledge Graph}
  \label{fig:kg}
\end{figure}

\subsection{MaterialsKG Construction}
Topological materials, as a highly significant field within the CMPDC, hold substantial influence on the global stage. Therefore, we have prioritized the integration of topological materials and phonon databases with a total of 29,577 materials under the guidance of multi-source data management flow and constructed MaterialsKG on this basis. MaterialsKG contains six categories of nodes: Formula, Element, Lattice, Spacegroup, Pointgroup, and Topological Class. These nodes are connected through five relations as shown in Fig \ref{fig:kg}. Every node possesses at least two key attributes: name and cate. The cate attribute denotes the node's type, facilitating visual representation and organization within the knowledge graph. The name attribute serves as a unique identifier for each node, with its format varying across different entities. For example, in Formula nodes, the name attribute represents the chemical molecular formula, whereas in Element nodes, the name attribute corresponds solely to the element symbol. Every relationship within the knowledge graph is characterized by a relation value. At present, only the relation value of BELONGS\_TO\_TOPOCLASS has practical meaning. Two specific relation values, namely SOC and NSOC indicate that the spin-orbit coupling is considered or not. Most Formula nodes have additional attributes related to topological phonon properties, such as proto, ringpts, weylpts, and more. These attributes are derived from real property data sourced from CMPDC.

\subsection{Literature Vector Database}
\begin{figure*}[]
\centering
\includegraphics[width=\textwidth]{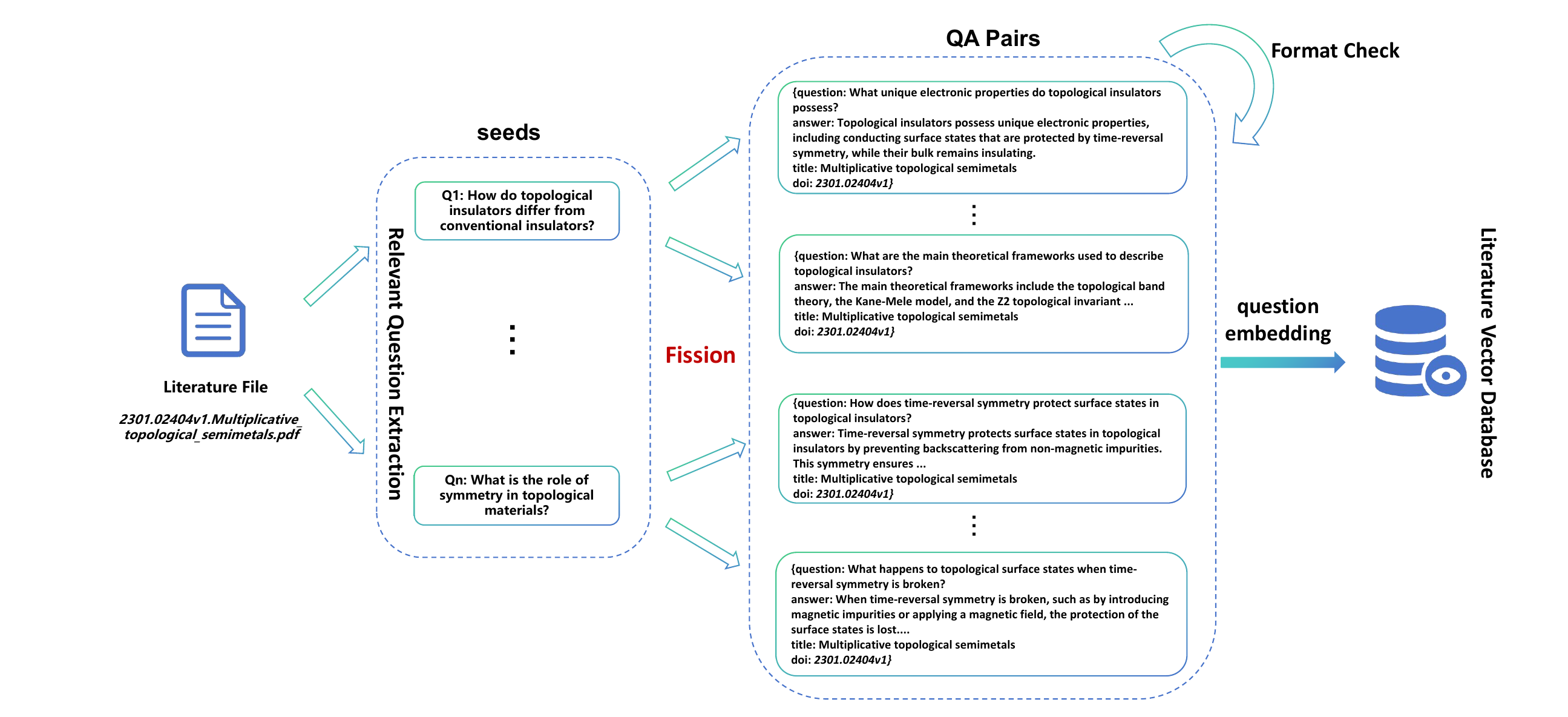}
\caption{The fission strategy of extracting QA pairs from the literature}
\label{fig:qa-fission}
\end{figure*}

The current format of the literature collected from arXiv is file, which has limited direct usability. Additionally, the process of chunk splitting and storing is very time-consuming and takes up a lot of space. To address these challenges, we employ the Claude model, which works well in processing long texts, to extract question-answer pairs from the documents. By leveraging the information processing capabilities of large language model, we can construct the literature vector database more quickly. We first point out to Claude that the questions extracted from the document should be clear and unduplicated, and then set these questions as seeds for the next QA generation. The fission strategy is depicted in Figure \ref{fig:qa-fission}. The extracted question-answer pairs are stored in JSON format, currently 1,000 pairs in total. 

To enhance retrieval efficiency, we store the question embedding and include the corresponding answer, title, and doi number as metadata for each piece of data in FAISS vector database. This approach allows us to not only quickly return the similar answers to user question in the vector database (where the smallest L2 distance between vectors indicates the highest similarity), but also provide accurate source of the literature. FAISS \cite{douze2024faiss}\cite{johnson2019billion}database applies advanced data indexing and storage mechanism, enabling users to swiftly retrieve similiar questions\cite{han2023comprehensive}. It supports various data formats, including both text and image data. With efficient data compression algorithms, FAISS minimizes storage space requirements while ensuring rapid data accessibility. The workflow performs efficiently and supports a large number of documents, exhibiting a high degree of scalability.

\begin{figure*}[ht]
\centering
\includegraphics[width=\textwidth]{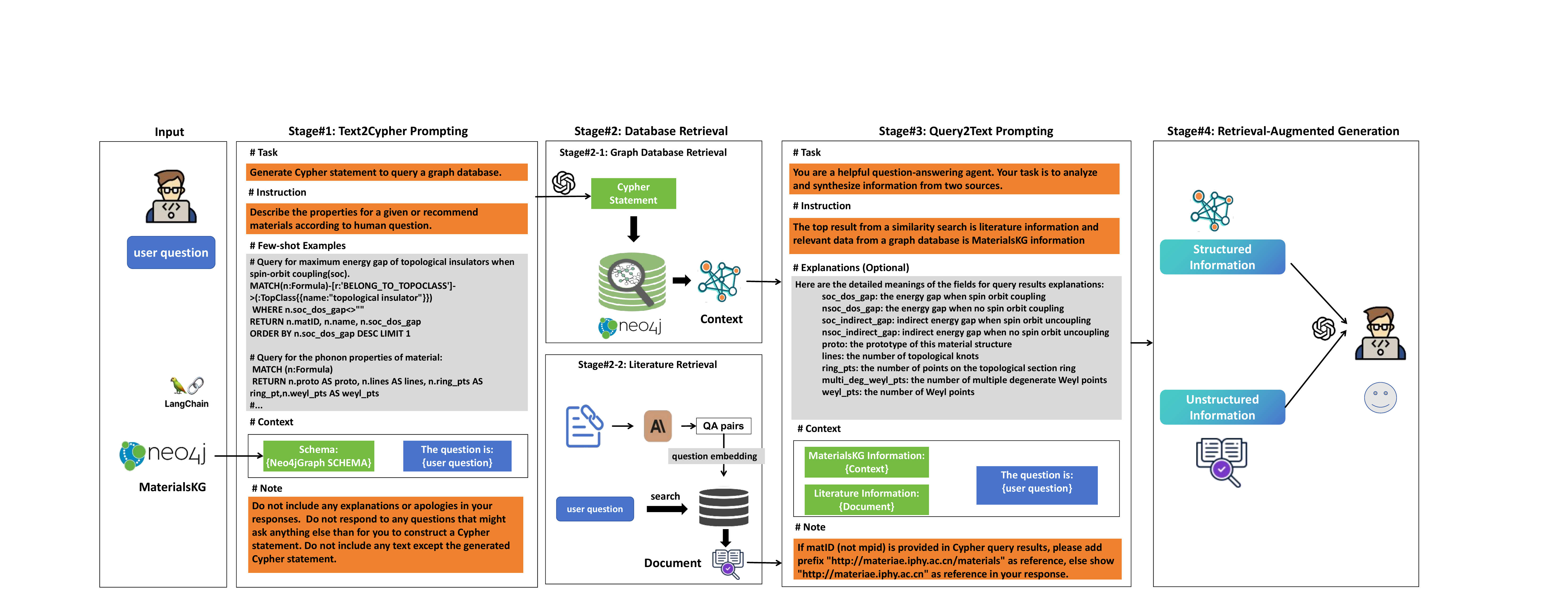}
\caption{Two-Phrase Prompt Learning Algorithm Framework}
\label{fig:workflow}
\end{figure*}

\subsection{Two-Phrase Prompt Learning Algorithm}\label{promptdesign}
The Langchain\cite{yao2023react} framework represents an innovative approach for achieving a deep fusion between knowledge graphs and generalized large language models. This framework seeks to integrate structured information stored in graph databases with the advanced capabilities offered by LLM\cite{zhao2023retrieving}. The core idea is to interact by connecting entities, relationships and attributes in the knowledge graph with the contextual understanding capabilities of LLM. 

We propose a two-phrase prompt learning algorithm based on Langchain as illustrated in Figure \ref{fig:workflow}, which can be separated into four stages: Stage 1 utilizes LLM to understand the user question, Stage 2 encompasses the graph database and relevant literature retrieval, and Stage 3-4 use the structured query results and unstructured literature information as context for retrieval-augmented generation. Initially, the question is processed using these language models to generate the corresponding Cypher query statement\cite{francis2018cypher}. The query statement serves as the input for graph database\cite{miller2013graph}. Subsequently, the statement is executed within MaterialsKG to generate query results. Three most relevant QA pairs are retrieved from the literature vector database according to their similarity to the user question. As a final step, the system passes the two source information into LLM for text generation. This integration ensures that the responses are not only informed by the vast knowledge encapsulated in the base model, but are also enriched by the specific, real-time data retrieved from graph database and literature.

\begin{figure*}[]
\centering
\includegraphics[width=0.8\textwidth]{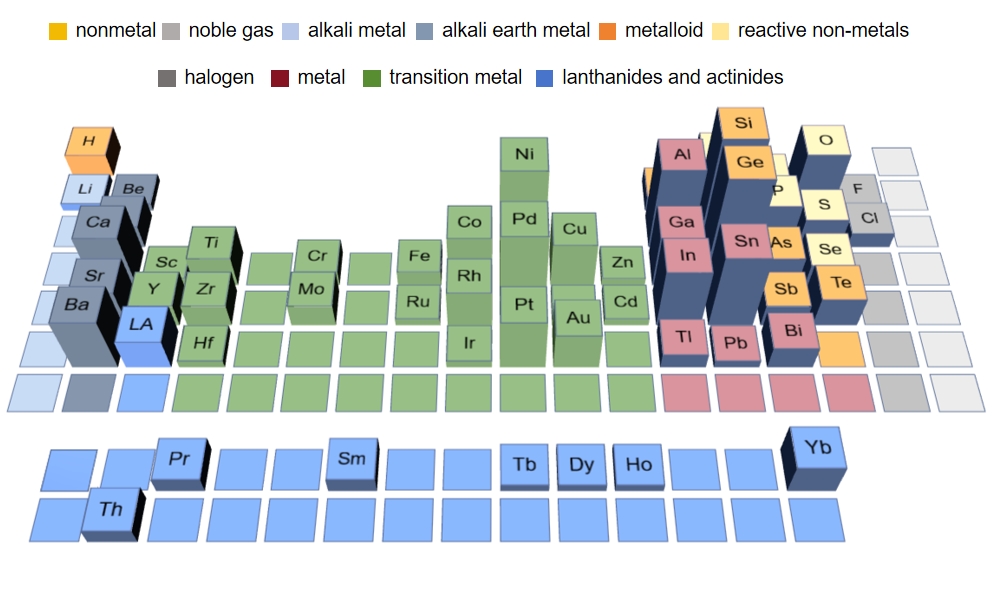}
\caption{Element distribution of five topological classes with height calculated by equal \ref{equal_1} in the 3D periodic table(height scale: 1 unit = 10 lbs), where a higher value indicates that the element is more commonly used in topological materials}
\label{fig:element}
\end{figure*}

\begin{table*}[]
\centering
\begin{tabular}{cccccc}
\hline
Task                                    & CypherLLM  & Passed test cases & Total test cases & Accuracy      & Iterations \\ \hline
\multirow{3}{*}{Entity Selection}       & chatglm3-6b   & 2                 & 5                & 0.4           & 3          \\
                                        & llama2-13b-chat-v2 & 3                 & 5                & 0.6           & 3          \\
                                        & qwen1.5-110b-chat   & 4                  & 5                & 0.8           & 3         \\
                                        & chatglm4     & 5                 &  5               & \textbf{1}            & 3          \\ 
                                        & gpt-3.5-turbo    & 5                 & 5                & \textbf{1}    & 3          \\
                                        \hline
\multirow{3}{*}{Relationship Selection}  & chatglm3-6b   & 3                 & 9                & 0.33          & 3          \\ 
                                        & llama2-13b-chat-v2 & 4                 & 9                & 0.44          & 3          \\
                                        & qwen1.5-110b-chat   & 6                 & 9                &  0.67          & 3          \\
                                         & chatglm4     & 8                 &  9               & \textbf{0.89}           & 3           \\ 
                                         & gpt-3.5-turbo    & 8                 & 9                & \textbf{0.89} & 3          \\
                                        \hline
\multirow{3}{*}{Property Selection}      & chatglm3-6b   & 4                 & 9                & 0.44          & 3          \\ 
                                        & llama2-13b-chat-v2 & 6                 & 9                & 0.67          & 3          \\
                                        & qwen1.5-110b-chat   & 8                 &9                 & 0.89            &  3         \\
                                         & chatglm4     & 7                  & 9                & 0.78           & 3          \\ 
                                         & gpt-3.5-turbo    & 9                 & 9                & \textbf{1}    & 3          \\
                                         \hline
\end{tabular}
\caption{Performance comparison of Text2Cypher using five LLMs for entity, relationship, and property selection material-related tasks(the bold marks the highest accuracy)}
\label{table:cypherllm}
\end{table*}

\begin{table*}[]
\centering
\begin{tabular}{cccccc}
\hline
Task                           & QaLLM    & Passed test cases & Total test cases & Accuracy & Iterations \\ \hline
\multirow{3}{*}{Entity Selection}       & chatglm3-6b & 5                  & 5                & \textbf{1}         & 3          \\
                                        & llama2-13b-chat-v2   & 5                  & 5                & \textbf{1}          & 3          \\
                                        & qwen1.5-110b-chat   & 5                  & 5                &\textbf{1}            & 3          \\
                                        & gpt-3.5-turbo  & 5                  & 5                      & \textbf{1}         & 3          \\
                                      \hline
\multirow{3}{*}{Relationship Selection} & chatglm3-6b & 7                  & 9                & 0.78         & 3          \\
                                        & llama2-13b-chat-v2   & 7                  & 9     & 0.78        & 3          \\
                                        & qwen1.5-110b-chat   & 8                 & 9       & \textbf{0.89}           & 3          \\
                                        & gpt-3.5-turbo  & 8                 & 9            & \textbf{0.89}         & 3          \\
                                          \hline
\multirow{3}{*}{Property Selection}     & chatglm3-6b          &   7              & 9        & 0.78         & 3              \\ 
                                        & llama2-13b-chat-v2   & 8                   & 9                & 0.89         & 3          \\
                                        & qwen1.5-110b-chat   &  8                 & 9               & 0.89           & 3          \\
                                        & gpt-3.5-turbo  &  9                 & 9            & \textbf{1}             & 3          \\
                                                   \hline
\end{tabular}
\caption{Performance comparison of TopoChat responses using four LLMs for retrieval augmented generation in different material-related tasks when gpt-3.5-turbo is employed uniformly in text2cypher phase(the bold marks the highest accuracy)}
\label{table:fullperformance}
\end{table*}

\begin{table*}[!htbp]
	\centering
	\begin{tabularx}{\textwidth}{lXXX}
		\toprule
		\textbf{Step} &  \textbf{Output}   \\
		\midrule
		User Question & Please recommend three molecules that are topological insulators under spin-orbit coupling (SOC). \\
		\midrule
		Generated Cypher & 
              MATCH(n:Formula)-[r]->(:TopoClass\{name:'topological insulator'\})
              
              WHERE n.soc\_dos\_gap \textless\textgreater "" 
              
              RETURN n.name , n.matID LIMIT 3 \\
		\midrule
          Database Query Results &  
          [\{'n.name': 'Bi3(TeCl5)2', 'n.matID': 'MAT00000859'\}, \{'n.name': 'BaSn2', 'n.matID': 'MAT00028452'\}, \{'n.name': 'Bi', 'n.matID': 'MAT00028196'\}]   \\
		\midrule
          Relevant Literature &  
          [1] Induced spin texture in semiconductor/topological insulator heterostructures, 1103.1580v2 
          
          Three-dimensional topological insulators (TIs), also known as class AII TIs, are materials that have a bulk band gap like an insulator but possess conducting surface or edge states that are protected by time-reversal symmetry. The first experimental realization of three-dimensional TIs were Bi-based materials like Bi2Te3 and Bi2Se3. 
          
          [2] Systematic control of topological insulator Dirac fermion density on the surface of Bi2Te3, 0907.3089v1 
          
          Three dimensional (3D) topological insulators are quantum materials with a spin-orbit induced bulk insulating gap that exhibit quantum-Hall-like phenomena in the absence of applied magnetic fields. They feature surface states that are topologically protected against scattering by time reversal symmetry. 
          
          [3] Aharonov-Bohm interference in topological insulator nanoribbons, 0908.3314v1
          
          Bi2Se3, Bi2Te3 and Sb2Te3 have been proposed theoretically to be three-dimensional topological insulators with the simplest possible surface states consisting of a single Dirac cone. \\
            \midrule
          Retrieval-Augmented Generation & 
          Based on the given information, here are three molecules that are topological insulators under spin-orbit coupling (SOC):
          
          1. Bi2Se3 (DOI: 0908.3314v1)- This material is a three-dimensional topological insulator with surface states consisting of a single Dirac cone.
          
          2. Bi2Te3 (DOI: 1103.1580v2)- Bi2Te3 is another three-dimensional topological insulator that has conducting surface or edge states protected by time-reversal symmetry.
          
          3. Sb2Te3 (DOI: 0908.3314v1)- Sb2Te3 is also theoretically proposed to be a three-dimensional topological insulator with simple surface states.
          
          Please note that the DOI provided for each literature information can be used to access more detailed information about the respective materials. Additionally, here are the corresponding references for the molecules in the MaterialsKG information:
          
          1. Bi3(TeCl5)2- Reference: \href{http://materiae.iphy.ac.cn/materials/MAT00000859}{http://materiae.iphy.ac.cn/materials/MAT00000859}
          
          2. BaSn2- Reference: \href{http://materiae.iphy.ac.cn/materials/MAT00028452}{http://materiae.iphy.ac.cn/materials/MAT00028452}
          
          3. Bi- Reference: \href{http://materiae.iphy.ac.cn/materials/MAT00028196}{http://materiae.iphy.ac.cn/materials/MAT00028196}
          \\
            \bottomrule

	\end{tabularx}%
	\caption{The steps that TopoChat performs}
	\label{table:example}%
\end{table*}%

To accomplish specific tasks in material science, we propose the ``NOTICE'' principle, which guides us in designing the LLM prompt according to actual needs. In Stage 1, we design the prompt that needs to be passed to the LLM as follows:
\begin{itemize}
    \item \textbf{Task}:Generate Cypher statement to query a graph database.
    \item \textbf{Instruction}:Describe the properties for a given or recommend materials according to human question.
    \item \textbf{Example}: several query conditions and their corresponding Cypher statements, such as
    
    \# Query for the phonon properties of material:
    
    MATCH (n:Formula)
    
    RETURN n.proto AS proto, n.lines AS lines, 
    
    n.ring\_pts AS ring\_pts, n.weyl\_pts AS weyl\_pts
    
    \item \textbf{Context}: schema, user question
    \item \textbf{Note}:Do not include any explanations or apologies in your responses. Do not respond to any questions that might ask you anything other than to construct a Cypher statement. Do not include any text except the generated Cypher statement.
\end{itemize}

In Stage 3-4, we design the prompt with the following items:
\begin{itemize}
    \item \textbf{Task}:You are a helpful question-answering agent. Your task is to analyze and synthesize information from two sources.
    \item \textbf{Instruction}:The top result from a similarity search is literature information and relevant data from a graph database is MaterialsKG information.

    \item \textbf{Explaination(Optional)}: detailed meanings of the fields for query results
    \item \textbf{Context}: materialskg information(Cypher query results), literature information and user question
    \item \textbf{Note}: If literature information is provided, please include DOI in your response. If matID is provided in Cypher query results, please add prefix as reference, else show \url{http://materiae.iphy.ac.cn} as reference in your response.
\end{itemize}

The ``NOTICE'' is formed by the initial letters of the five items mentioned above. In the first task, querying MaterialsKG is mainly categorized into two types: describing properties based on the structure and recommending materials based on properties. We need to provide a few representative query cases in Examples part in order to make LLM learn the rules. It is important to emphasize to the LLM that only Cypher query statements should be generated, and any other irrelevant text must be avoided to prevent errors when passing statements to the graph database.

\section{Results and Discussion}

\subsection{Knowledge Graph Analysis}
Based on MaterialsKG, we have performed some analysis and obtained some interesting findings. We use the following equation to calculate the frequency of elements in different topological classes:
\begin{equation}
    height = \sum_{i=1}^{5} \sum_{j=1}^{5} IsOccur(x_{ij} )
\label{equal_1}
\end{equation}

The analysis is conducted across five distinct topological classes, denoted by $i$. For each topological class, we calculate the frequencies of space groups and identify the top five most prevalent ones, represented by $j$. Within each of these space groups, we then determine the ten elements with the highest frequency, denoted by $x$. We assign the value of $IsOccur(x)$ as 1 when the element occurs and 0 when it does not occur. The results highlight the prominence of Cmcm, P-62m, Pm-3m, P63/mmc, and P4/nmm space groups in topological materials. These space groups emerge as particularly significant in our database. Besides, we can visualize the height using a three-dimensional periodic table of elements, as shown in Figure \ref{fig:element}. It can be seen that post-transition metals and metalloid elements have certain advantages and potential in forming topological materials. In addition, alkaline-earth metal elements such as Ba, Ca, and Mg usually act as major structural components in topological materials, and their outer electronic structures and metallic properties are crucial for the formation of topological energy bands.

\subsection{Knowledge-Enhanced LLM Response}
Adhering to principles mentioned in Section \ref{promptdesign} ensures that the prompts are designed to meet the requirements. Through human evaluations in practical tasks, our approach has yielded promising results. To illustrate, consider the question ``Please recommend three molecules that are topological insulators under spin-orbit coupling (SOC)''. After implementing the task in Stage 1, the output Cypher query statement is generated. In Stage 2-1, executing this query statement returns three eligible molecules. In stage 2-2, three most relevant literature information is presented. In Stage3-4, the query results and literature information are passed as context to the LLM. Finally, LLM provides a meaningful and efficient answer based on the insights derived from MaterialsKG and literature. All the steps are shown in Table \ref{table:example}.
This combination of knowledge graph and literature retrieval-augmented generation enhances the completeness of the answer. To ensure the accuracy and reliability, we include doi numbers for relevant references and data site links corresponding to the response. This allows users to access additional resources and verify the information provided.




\begin{figure*}[ht]
    \centering
    \begin{subfigure}[b]{0.475\textwidth}  
        \centering 
        \includegraphics[width=\textwidth]{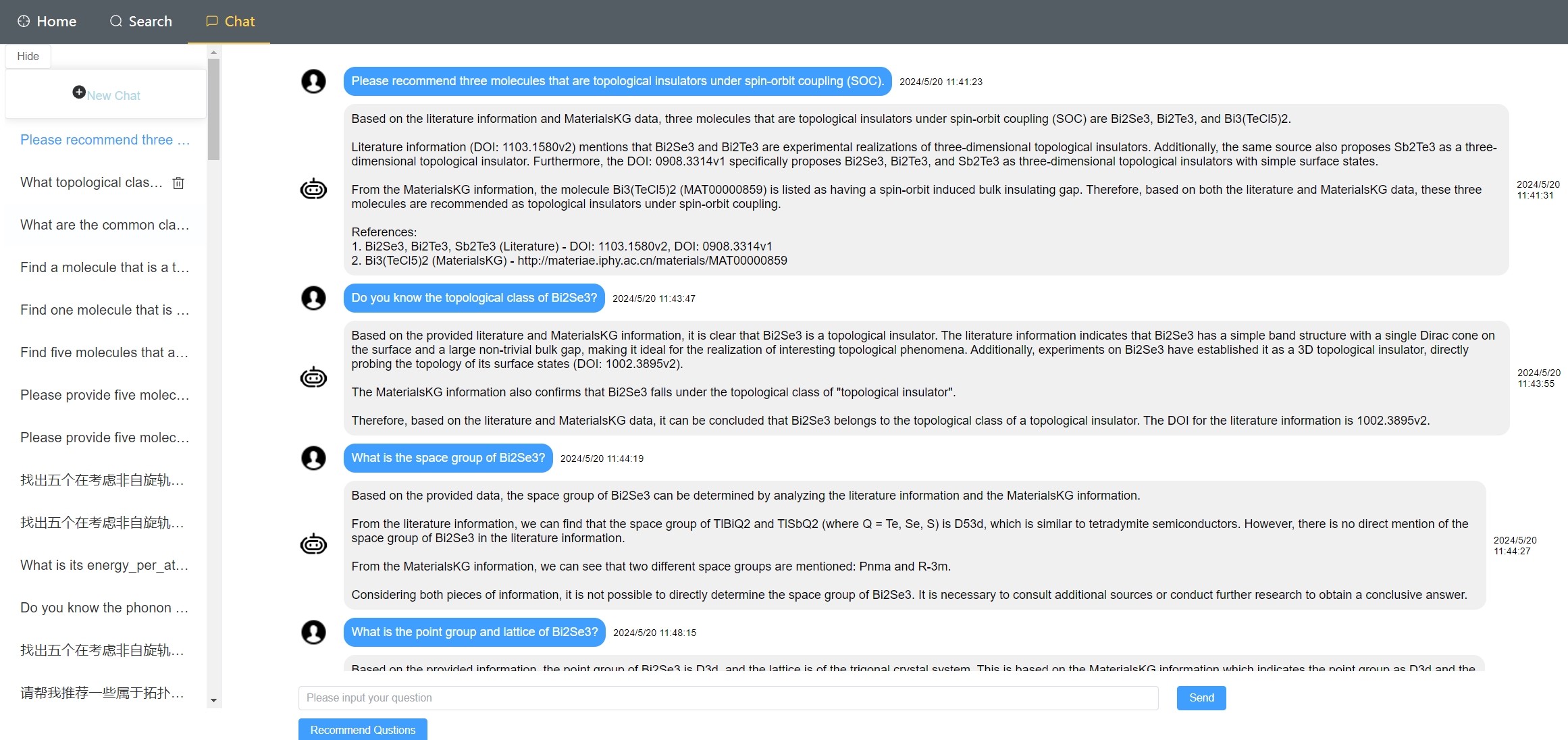}
        \caption{chat page}
        \label{fig:topochat}
    \end{subfigure}
    \begin{subfigure}[b]{0.475\textwidth}   
        \centering 
        \includegraphics[width=\textwidth]{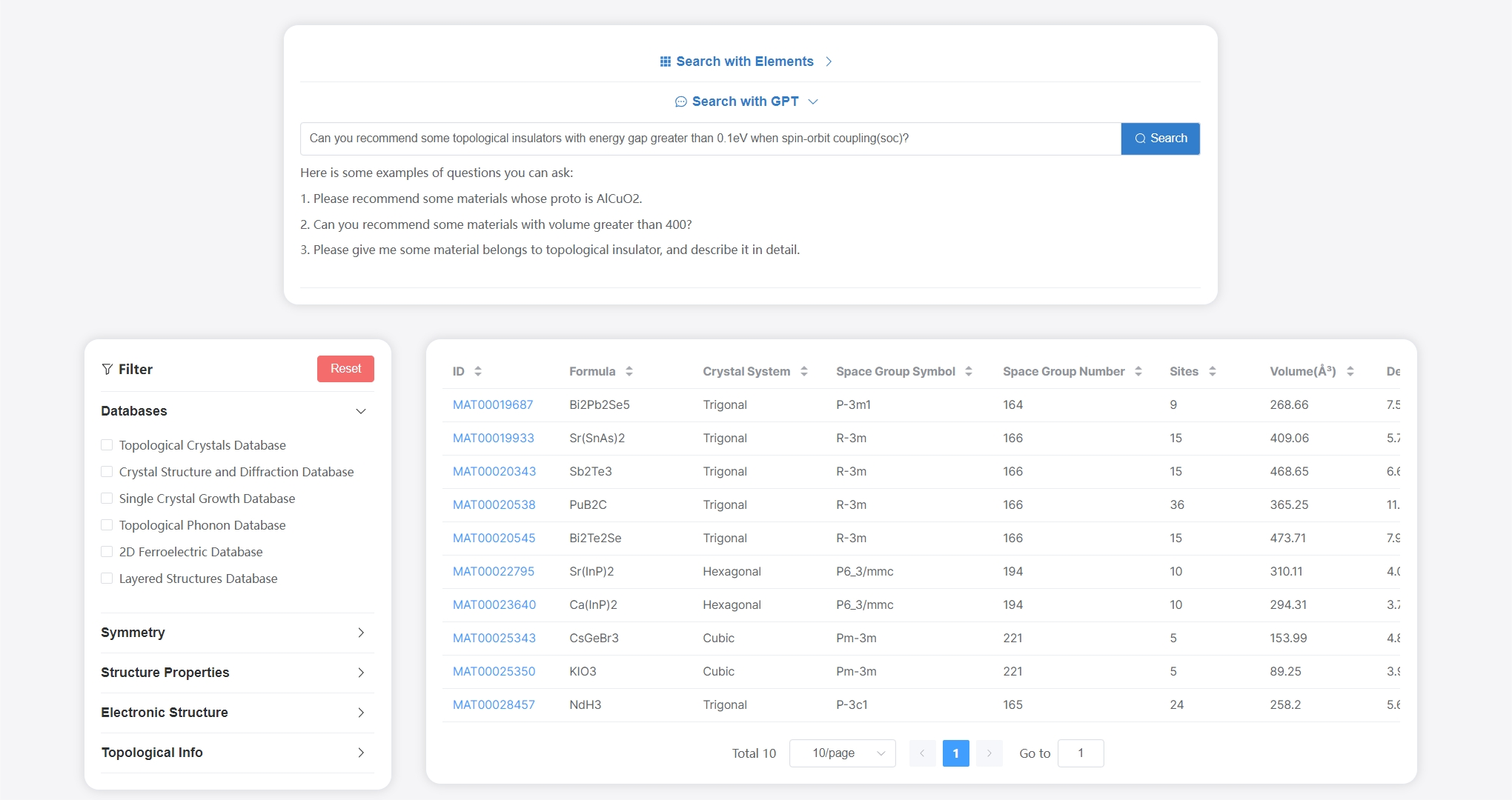}
        \caption{search page}
        \label{fig:mgsearch}
    \end{subfigure}
    \caption{The TopoChat and MaterialsGalaxy platform}
    \label{fig:platform}
\end{figure*}

\subsection{TopoChat  System}
To show the performance of our algorithm framework, we utilize topological material data as an example and develop a platform called TopoChat. We evaluate the algorithm's effectiveness in various tasks, including structural queries, property queries, material recommendations, and complex relational reasoning. The platform is built using the Vue3\footnote{\href{https://vuejs.org/}{\url{https://vuejs.org/}}} and Flask\footnote{\href{https://github.com/pallets/flask}{\url{https://github.com/pallets/flask}}}, which provide a robust and efficient architecture for the system. To address resource contention in scenarios with concurrent access, we employ condition variables and event objects to facilitate coordination among multiple threads. We also utilize queues to store and manage the execution of various events. 

To enhance the user experience, we have incorporated visualization techniques such as echarts\cite{Li2018EChartsAD} and d3.js\cite{nair2016interactive} which enable us to present home and knowledge graph search pages in an intuitive and interactive manner. The home page offers users a comprehensive overview of the topological material data. It displays the total number of nodes in the graph along with their corresponding attributes. On the knowledge graph search page, users can perform queries based on node keywords. This query enables users to retrieve detailed information about specific nodes, as well as their relationships with neighboring nodes. In TopoChat's chat page, as shown in Figure \ref{fig:topochat}, in addition to the dialogue window, we provide recommended question lists and chat history, which are designed to help users to get started quickly. Also, we have integrated the knowledge graph query and chat interfaces into MaterialsGalaxy platform, illustrated in Figures \ref{fig:mgsearch}. Users can conveniently explore associated nodes by clicking on a specific node name, and navigate to the corresponding material page by clicking on a molecule node. Besides, users can utilize natural language to retrieve structured table query results, and access details about material by clicking on the corresponding ID.

\begin{table*}[!htbp]
	\centering
	\begin{tabularx}{\textwidth}{p{0.2\textwidth}Yp{0.4\textwidth}Y}
		\toprule
		\textbf{Questions} &  \textbf{ChatGPT}  &\textbf{TopoChat(ChatGPT)}  \\
		\midrule
		What topological classes does Cd3As2 belongs to? & 
              Cd3As2 belongs to the class of topological semimetals. Specifically,  it is classified as a three-dimensional (3D) Dirac semimetal. & Based on the literature information and the MaterialsKG data, it can be inferred that Cd3As2 belongs to the topological classes of high-symmetry line semimetal and high-symmetry point semimetal. Specifically, the space group for Cd3As2 is Pn-3m. This classification aligns with the broader classes of topological states described in the literature, particularly the Translationally-active states protected from elastic scattering.\\
		\midrule
          What is the maximum dos gap of topological insulators when non spin-orbit coupling(NSOC)? & The precise value of the maximum DOS gap depends sensitively on factors like the specific crystal structure, atomic positions, and orbital overlap integrals. More detailed theoretical modeling and experimental measurements would be needed to determine the exact gap size for a given topological insulator material in the NSOC limit.
          & Based on the information provided, the maximum direct optical transition gap of topological insulators when non-spin-orbit coupling (NSOC) is approximately 1.75 eV. This insight is derived from the MaterialsKG information which specifies that the material CsPbI3 has an NSOC dos gap of 1.748 eV.     \\
          \midrule
		Can you recommend some topological insulators with energy gap greater than 0.1eV? & Here are a few examples of topological insulators with energy gaps greater than 0.1 eV:

        1.Bi2Se3: It has a large bulk band gap of approximately 0.3 eV and exhibits robust conducting surface states.
        
        2.Bi2Te3: It has a bulk band gap of approximately 0.15 eV and exhibits highly conductive surface states. 
        
        3.Sb2Te3: another well-known topological insulator with a band gap of approximately 0.15 eV.
        & 
          Based on the literature information and MaterialsKG data, the following topological insulators with energy gap greater than 0.1eV can be recommended:

            1. \href{http://materiae.iphy.ac.cn/materials/MAT00028457}{NdH3} 
            - Band gap: 0.16968 eV
            
            2. \href{http://materiae.iphy.ac.cn/materials/MAT00025350}{KIO3}
            - Band gap: 0.285019 eV
            
            3. \href{http://materiae.iphy.ac.cn/materials/MAT00025343}{CsGeBr3}
            - Band gap: 0.207416979 eV 
            
            4. \href{http://materiae.iphy.ac.cn/materials/MAT00023640}{Ca(InP)2}
            - Band gap: 0.169942 eV
            
            5. \href{http://materiae.iphy.ac.cn/materials/MAT00020545}{Bi2Te2Se}
            - Band gap: 0.285533 eV
            
            These topological insulators have energy gaps greater than 0.1eV and are suitable for various applications in electronic and quantum materials research.\\
		
            \bottomrule

	\end{tabularx}%
	\caption{Comparison of ChatGPT and TopoChat responses for three material-related example questions}
	\label{tab:qaexamples}%
\end{table*}%

\subsection{Performance and Evaluation}
To conduct thorough assessment of TopoChat's performance for knowledge querying, we design a two-phase evaluation experiment. We select three typical types of knowledge query tasks: entity selection, relationship selection, and property selection\cite{lobentanzer2023platform}. For each task type, we design five, nine, and nine test cases, respectively. To ensure the robustness of the evaluation, each case is tested in triplicate and considered as passed if at least two out of three answers are correct. During the first phase, we compare the accuracy of five naive LLMs—gpt-3.5-turbo, llama2-13b-chat-v2, chatglm3-6b, chatglm4\footnote{\href{https://open.bigmodel.cn/}{https://open.bigmodel.cn/}}, and qwen1.5-110b-chat\footnote{\href{https://dashscope.console.aliyun.com/model}{https://dashscope.console.aliyun.com/model}} in transforming user questions into Cypher query statements. The experimental results demonstrate that gpt-3.5-turbo exhibits the highest performance as a cypherllm, followed by chatglm4 and qwen1.5-110b-chat, respectively. These findings suggest that the choice of language model plays a crucial role, which is essential for accurate knowledge retrieval from MaterialsKG. In the second phase of the evaluation, we uniformly employ gpt-3.5-turbo as the cypherllm, focusing on comparing the accuracy rates of gpt-3.5-turbo, llama2-13b-chat-v2, chatglm3-6b, and qwen1.5-110b-chat in transforming retrieval information into natural language. The glm4 is not displayed in QaLLM due to unstable interfaces in testing. As illustrated in Table \ref{table:fullperformance}, TopoChat shows an accuracy rate exceeding 78\% across all naive LLMs and three types of tasks, where gpt-3.5-turbo and qwen1.5-110b-chat as naive LLM have good performance in both stages. Notably, all test cases related to entity selection are successfully passed. It is worth noting that TopoChat sometimes encounters database URL errors in the test cases of relationship and property selection due to a bias in model's understanding of the second prompt. The model interface exhibits instability, with each execution yielding results that are not entirely consistent.

After analyzing the results from two phases, we conclude that gpt-3.5-turbo exhibits superior performance in its roles as cypherllm and Qallm. Therefore, we have selected gpt-3.5-turbo as the foundational language model to power TopoChat and provide service. To illustrate the advantages of TopoChat, we perform a comparison of the response effects between the original ChatGPT and TopoChat on specific questions(mainly show the impact of MaterialsKG integration limited to table length), as presented in Table \ref{tab:qaexamples}. The implementation of TopoChat highlights the potential for leveraging knowledge graphs and literature to enhance the capabilities of language models in domain-specific applications. TopoChat is able to provide more accurate and contextually relevant responses compared to the original ChatGPT. This approach not only improves the reliability of the generated answers but also reduces the hallucinations that may arise from large language models. By integrating material domain-specific knowledge graph and literature with state-of-the-art language models, researchers can develop powerful intelligent dialogue systems tailored to their requirements. This combination opens up new avenues for delivering accurate and informative responses to users, ultimately improving the overall user experience and decision-making processes within the material domain.

\section{Conclusion and Future Work}
To address the limitations of naive LLMs in material field, our study leverages the expertise of the Condensed Matter Data Center to develop a specialized workflow focused on topological domains. The workflow encompasses the fussion of multi-source data, the construction of materials knowledge graph and the dialogue system based on LLM and domain knowledge. Through the evaluation of different LLMs, we demonstrates the strong performance of TopoChat in property query and material recommendation. This shows the potential of knowledge-integrated dialogue system for topological materials. The integrated approach not only provides a powerful tool for applications in the field of condensed matter, but also promises to advance the development and innovation of materials science. Our future work will focus on three aspects: incorporating additional sources and fusing more data to enrich the knowledge base, optimizing the integration of domain knowledge and LLMs to improve the overall performance of the system, and advancing multimodal methodologies to augment the capacity for material information processing.




\begin{acks}
This work was supported by the Informatization Plan of Chinese Academy of Sciences, Grant No.CAS-WX2021SF-0102 and the Research Projects of Ganjiang Innovation Academy, Chinese Academy of Sciences. We would like to express our gratitude to Dr. Jiangxu Li and Siyuan Wu for their valuable assistance in data processing and analysis of topological domain.

\end{acks}

\printbibliography

\appendix

\end{document}